\documentclass{article}

\usepackage{PRIMEarxiv}

\usepackage[utf8]{inputenc} 
\usepackage[T1]{fontenc}    
\usepackage{hyperref}       
\usepackage{url}            
\usepackage{booktabs}       
\usepackage{amsfonts}       
\usepackage{nicefrac}       
\usepackage{microtype}      
\usepackage{lipsum}
\usepackage{fancyhdr}       
\usepackage{graphicx}       
\usepackage{subcaption}     
\usepackage{amsmath}        
\usepackage{multirow}       
\DeclareMathOperator*{\argmax}{argmax}

\graphicspath{{media/}}     

\title{TOP-Former: A Multi-Agent Transformer Approach for the Team Orienteering Problem
}

\author{
    Daniel Fuertes,
    Carlos R. del-Blanco,
    Fernando Jaureguizar,
    Narciso García \\
    Grupo de Tratamiento de Imágenes (GTI),
    Information Processing and Telecommunications Center, \\
    ETSI Telecomunicación,
    Universidad Politécnica de Madrid,
    28040 Madrid,
    Spain \\
    \texttt{
        \{d.fcoiras,
        carlosrob.delblanco,
        fernando.jaureguizar,
        narciso.garcia\}@upm.es
    }
}

\begin{document}
\maketitle

\begin{abstract}
Route planning for a fleet of vehicles is an important task in applications such as package delivery, surveillance, or transportation, often integrated within larger Intelligent Transportation Systems (ITS). This problem is commonly formulated as a Vehicle Routing Problem (VRP) known as the Team Orienteering Problem (TOP). Existing solvers for this problem primarily rely on either linear programming, which provides accurate solutions but requires computation times that grow with the size of the problem, or heuristic methods, which typically find suboptimal solutions in a shorter time. In this paper, we introduce TOP-Former, a multi-agent route planning neural network designed to efficiently and accurately solve the Team Orienteering Problem. The proposed algorithm is based on a centralized Transformer neural network capable of learning to encode the scenario (modeled as a graph) and analyze the complete context of all agents to deliver fast, precise, and collaborative solutions. Unlike other neural network-based approaches that adopt a more local perspective, TOP-Former is trained to understand the global situation of the vehicle fleet and generate solutions that maximize long-term expected returns. Extensive experiments demonstrate that the presented system outperforms most state-of-the-art methods in terms of both accuracy and computation speed.
\end{abstract}


\section{Introduction}
\label{sec:intro}

The inclusion of vehicles like Unmanned Aerial Vehicles (UAV), Unmanned Ground Vehicles (UGV), or autonomous cars in applications like transportation \cite{Juan2020,Bono2021,Che2024}, 
package delivery \cite{Raivi2023,Loquercio2018}, search and rescue \cite{Wan2024,Yanmaz2023,Li2023}, and surveillance \cite{Dilshad2020,Fuertes2022} has many advantages, from maximizing the performance up to reducing the operational cost. These applications are often part of larger intelligent transportation systems (ITS), which aim to optimize traffic flow, safety, and logistics using advanced communication and computation technologies. Such systems typically rely on multiple vehicles to address the challenge of visiting a massive number of locations. A key task that arises in the management of vehicle fleets is the automation of route planning, especially when they must cooperate to achieve high operational performance. Multi-agent route planning algorithms must consider a global view of the problem to balance the workload among agents and thereby maximize efficiency. Additionally, the possible solutions must meet constraints related to energy consumption (battery/fuel), introducing time restrictions for the routes of the agents. Traditionally, the problem of generating routes for multiple vehicles under time constraints has been formulated as the Team Orienteering Problem (TOP).

The TOP was first presented in \cite{Chao1996} as a Vehicle Routing Problem (VRP) \cite{Bogyrbayeva2024}. More specifically, it is a multi-agent variant of the Orienteering Problem (OP) \cite{Golden1987}, where an agent receives a reward after visiting a region, and the objective is to maximize the total reward collected while returning to the depot (the end depot can be the same as or different from the start depot) within a given time limit (due to the aforementioned limitations of energy consumption). Compared to OP, which is already a NP-hard (Non-deterministic Polynomial-time hard) problem, the TOP is much more complex to solve due to the existence of multiple agents that should be coordinated.

Similarly to the OP, linear programming methods have been proposed to solve the TOP problem \cite{Bianchessi2018,Kobeaga2024,Sundar2022,Zamorano2017}, but they are usually very computationally demanding and cannot infer solutions for real-time applications that additionally can require the use of hardware with limited on-board computational capabilities (such as in UAVs). Other works have attempted to reduce the complexity of TOP by introducing heuristics in exchange of some performance loss. For example, dividing the scenario into mutually exclusive zones for each agent \cite{Sung2020,Pedersen2022}. Thus, the TOP is cast into many individual OPs, delegating the task of avoiding that two agents visit the same region to the initial scenario partition strategy. To compensate in some degree the lost in performance due to these heuristic simplifications, a region-sharing cooperative strategy is proposed in \cite{Fuertes2023}, relaxing the revisiting constraint of the TOP and allowing that more than one agent can visit a region under specific conditions. However, the global performance of previous algorithms is suboptimal since each agent does not take the decisions of the others into account. Also, the cooperation among agents is limited, with at most some mechanisms devised to recover part of the lost performance, such as \cite{Fuertes2023} with its region-sharing strategy.

In this paper, we propose TOP-Former, a novel multi-agent planning neural network that efficiently tackles the TOP by directly addressing the problem without relying on heuristics, thus achieving superior routing solutions. TOP-Former leverages a Transformer architecture \cite{Vaswani2017} that is trained with Deep Reinforcement Learning (DRL) to effectively learn to search for multiple cooperative and accurate routes for the TOP. The Transformer comprises a centralized Encoder-Decoder structure evolved from the decentralized solution of \cite{Fuertes2023} to consider the state of every agent before making predictions. The Encoder of TOP-Former projects the available information about the graph of nodes/regions to be visited into a graph embedding, and the situation of the agents (at every time step) into several state embeddings. On the other hand, the Decoder operates in an autoregressive manner and combines the encoded graph embedding and the state embedding of every agent to contextualize their situation at each time step (particularly with regard to their position and elapsed time), hence, improving the understanding of the network and increasing the quality of the solutions. This is contrary to \cite{Fuertes2023}, where the problem is decomposed into multiple sub-problems to analyze only the state of individual agents and solve an individual OP for each agent.

Therefore, the primary contribution of this paper is the presentation of TOP-Former, a centralized solution for the TOP that enables efficient route prediction and multi-agent coordination. In modern transportation networks, optimizing the movement of multiple autonomous vehicles is crucial for enhancing efficiency, reducing operational costs, and improving service quality. TOP-Former tackles this challenge by considering the global context of the entire fleet of agents to infer simultaneous, intelligent, and cooperative routes. Unlike traditional heuristics or optimization methods that struggle with scalability and real-time execution, TOP-Former utilizes a Transformer network to rapidly generate high-quality routing solutions, making it suitable for real-time applications. TOP-Former is trained using DRL simulations, which allow random synthetic scenario generation. To validate its effectiveness in intelligent transportation applications, TOP-Former is evaluated on the VDRPMDPC (Van Drone Routing Problem with Multiple Delivery Points and Cooperation) \cite{Athanasiadis2023} dataset, which is specifically designed for package delivery scenarios.

The organization of this paper is as follows. First, some relevant TOP solvers from the state of the art are exposed in Section \ref{sec:sota}. Next, in Section \ref{sec:problem}, the TOP formulation is detailed. Section \ref{sec:system} is dedicated to explaining the architecture of TOP-Former. The results of the experiments and comparisons with other algorithms are described in Section \ref{sec:results}. Finally, Section \ref{sec:conclusion} contains all the conclusions extracted from this work.

\section{Related works}
\label{sec:sota}

The TOP poses a significant complexity due to its NP-Hard nature. Traditionally, linear programming methods employing cutting-planes have been utilized to tackle this problem. Cutting-plane algorithms involve the application of linear inequalities (cuts) to refine the feasible set of solutions for the TOP. Notable works employing cutting-planes for the TOP include \cite{Bianchessi2018,Kobeaga2024}, which utilize the Branch-and-Cut method, as well as \cite{Sundar2022,Zamorano2017}, which employ the Branch-and-Prize algorithm. In addition to cutting-plane methods, there exist linear programming-based solvers tailored for general optimization problems that can be adapted, with certain assumptions, for the TOP. However, these solvers often focus more on the single-agent case. Examples include OR-Tools \cite{Ortools2024}, an open-source software developed by Google specifically dedicated to solving vehicle routing problems; Gurobi \cite{Gurobi2024}, a commercial solver capable of yielding highly accurate results but may face intractability issues with large TOP instances; and CPLEX \cite{Cplex2022}, another commercial solver similar to Gurobi that belongs to IBM.

The main drawback of previous linear programming methods lies in the extensive computational time required to attain highly accurate solutions, which can be problematic for real-world applications where fast decisions are expected to be taken by the agents. In response, alternative approaches leverage heuristic algorithms, prioritizing faster solutions over absolute accuracy. The following are notable examples of heuristic approaches applied to the TOP. In \cite{Ruizmeza2021}, a Greedy Randomised Adaptive Search Procedure (GRASP) is presented to face a variant of TOP that considers Time Windows (TOPTW) for visiting regions. In \cite{Ke2016}, a new heuristic method called Pareto mimic offers new solutions by mimicking a set of incumbent solutions updated iteratively according to Pareto dominance. A Monte Carlo Tree Search (MCTS) algorithm is designed in \cite{Shi2023} to solve a multi-agent variant of the OP. In \cite{Amarouche2020} a combination of Iterated Local Search (ILS) with some meta-heuristics is proposed. The Large Neighborhood Search (LNS) method of \cite{Shaw1998} is extended with a local search step in \cite{Hammami2020}. Furthermore, the benchmark introduced in \cite{Xiao2022} is remarkable since it adapts three popular heuristic algorithms, Particle Swarm Optimization (PSO), Ant Colony Optimization (ACO), and Genetic Algorithm (GA), for a variant of the TOP that relaxes the restriction of returning to the end depot, considering additional task-related temporal costs associated with each visited region. This set of heuristic solutions offers less accurate solutions than linear programming methods but decreases the computational time. Nonetheless, all previous heuristic methods are still too slow to be considered for real-time applications.

The recent emergence of DRL strategies to train neural networks has encouraged the appearance of works dedicated to solving VRPs. They have the potential to be much faster than previous methods, allowing the agents to operate in real-time. Inspired by Natural Language Processing (NLP) models based on Recurrent Neural Networks (RNN), they substitute word tokens for node tokens to sequentially predict the best route that solves the VRP. A good example are the Pointer Networks (PN) \cite{Vinyals2015}, which adapt the NLP model of \cite{Bahdanau2015} (combination of RNNs with Attention mechanisms) to face multiple problems such as the Travelling Salesman Problem (TSP). PNs were later extended by \cite{Bello2017}, adopting a Reinforcement Learning framework, and by \cite{Ma2020}, through the addition of a Graph Neural Network (GNN) to improve the performance of the model. Due to the sequential nature of PN that makes the training of new models too slow, \cite{Vaswani2017} proposed a Transformer neural network where the data is processed in parallel, improving both the accuracy of the model and the computation time. Notwithstanding that this Transformer model was developed for NLP tasks, it was later redesigned in \cite{Kool2019} to deal with VRPs.

Although the Transformer architecture presented in \cite{Kool2019} achieves high performance on different VRPs, it cannot solve multi-agent tasks. In fact, to the best of our knowledge, there are very few proposals that face multi-agent VRPs with DRL techniques due to their complex design. One remarkable work is \cite{Sykora2020}, which presents a Multi-Agent Routing Value Iteration Network (MARVIN) that works in a decentralized manner with several Attention-based LSTMs to control each vehicle in the fleet. Another is \cite{Sankaran2022}, which proposes a centralized Transformer, called Graph Attention Model for Multiple Agents (GAMMA). GAMMA is similar to \cite{Kool2019}, but with a different context embedding for encoding the agent information. More specifically, this context embedding switches among agents on each iterative step to make sequential predictions for all of them. From the mentioned works, MARVIN \cite{Sykora2020} performs well on small scenarios, but it does not scale well on large ones due to the nature of LSTMs. On the other hand, the solutions provided by GAMMA are suboptimal since the decision of each agent does not consider the state of the rest of the agents.

An additional multi-agent solution was proposed in \cite{Fuertes2023}, which consists in simplifying the problem by clustering the regions into different groups assigned to each problem and solving a variant of the OP for each agent. This variant of the OP considers a region-sharing strategy that alleviates the initial suboptimal clustering and promotes cooperation between agents. However, this approach cannot acquire a global understanding of all the context since each agent does not consider the decisions of the other agents.

To address the limitations of existing state-of-the-art methods, we propose TOP-Former, a direct and holistic solution to the TOP. Firstly, TOP-Former offers superior computational performance compared to both linear programming solvers and heuristic algorithms, enabling real-time operation. Secondly, we adopt a Transformer network for graph-attention predictions as the core of TOP-Former. This choice is supported by the demonstrated superiority of Transformer networks over other RNN and GNN-based solutions, as highlighted in \cite{Kool2019,Fuertes2022,Sankaran2022}. Thirdly, TOP-Former predicts multiple simultaneous routes while considering the global context of all agents, thereby enhancing the potential quality of solutions. This stands in contrast to sequential predictors that consider only the situation of one agent at a time. Moreover, by solving the TOP with a comprehensive understanding of the context of every agent, the task of learning to infer optimal predictions in the long run (which is particularly beneficial for large-scale scenarios) is facilitated to TOP-Former.

\section{Problem formulation}
\label{sec:problem}

\begin{figure*}[t]
    \centering
    \begin{tabular}{c|c}
        \begin{subfigure}{.45\textwidth}
            \centering
            \includegraphics[width=\columnwidth]{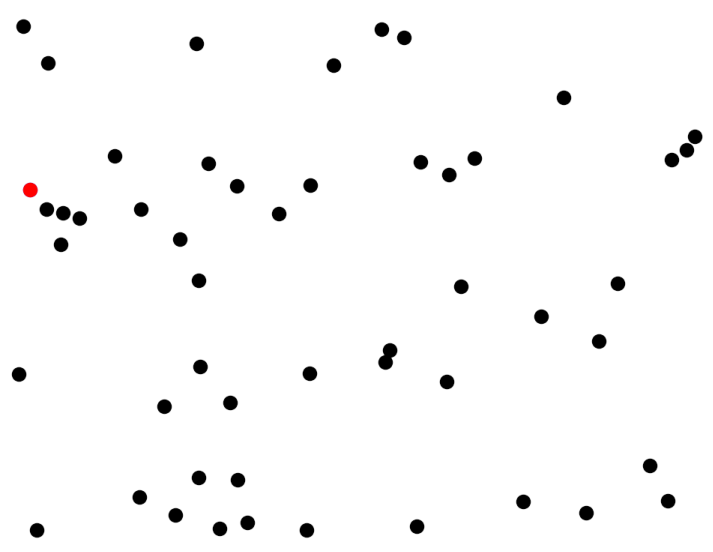}
            \caption{TOP scenario}
            \label{fig:scenario}
        \end{subfigure}
        &
        \begin{subfigure}{.45\textwidth}
            \centering
            \includegraphics[width=\columnwidth]{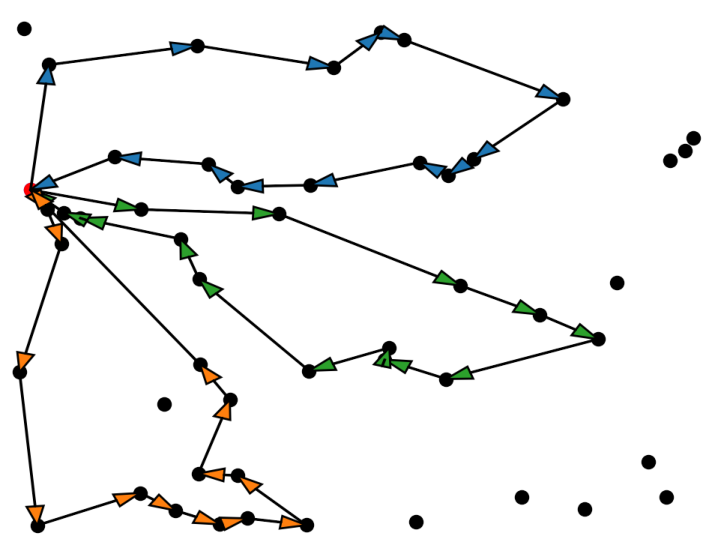}
            \caption{TOP solution}
            \label{fig:solution}
        \end{subfigure}
    \end{tabular}
\caption{Example of a TOP instance, containing: (a) a TOP scenario with a depot (red circle) and a set of regions (black circles); and (b) a solution for three agents (colored arrows). Notice that the TOP imposes a time limit to return to the end depot. Thus, it is not mandatory to visit all the regions.}
\label{fig:top_scenario}
\end{figure*}

Let $\mathcal{G}=(\mathcal{V}, \mathcal{E})$ denote a complete and undirected graph, where $\mathcal{V} = \{0, ..., n + 1\}$ represents a set of unique nodes that must be visited by a set of agents $\mathcal{A} = \{1, ..., m\}$, and $\mathcal{E} = \{(i, j): i, j \in \mathcal{V}, i \neq j\}$ denotes the set of symmetric edges connecting every pair of nodes in $\mathcal{V}$. Each node $i \in \mathcal{V}$ yields a reward $r_i$ when visited by an agent $a \in \mathcal{A}$. Nodes $i=0$ and $i=n+1$ have rewards $r_0=r_{n+1}=0$ as they correspond to the start and end depots, respectively, for all agents. Additionally, each edge $(i, j) \in \mathcal{E}$ has a non-negative cost $d_{ij}$ representing the distance between nodes $i$ and $j$, with $i \neq j, \forall \ i, j \in \mathcal{V}$. For simplicity, we assume Euclidean distances and constant agent speed $v^a=v, \forall a \in \mathcal{A}$.

The primary objective in the TOP is to identify a set of paths (i.e., a sequence of nodes), denoted as $\mathcal{P} = \{\rho^a \subset \mathcal{V}: a \in \mathcal{A}\}$, that maximizes the total reward collected starting from node $i=0$ and ending in node $i=n+1$ within a time limit $T$, as it is illustrated in Figure \ref{fig:top_scenario}. It can be expressed as follows:

\vspace*{-2mm}

\begin{flalign}
    \ \ \max_{\rho^a \subset \mathcal{P}} \
        \sum_{a = 1}^{m} &
            \sum_{l = 1}^{|\rho^a|}
                r_{\rho^a_l} \phi_{\rho^a_{l-1} \rho^a_l},
    & \label{eq:goalFunc} \\
    \textrm{s.t.} \quad
        & \sum_{j = 1}^{n + 1}
            \phi_{0j} = m
        & \label{eq:constraint1} \\
        & \sum_{i = 0}^{n}
            \phi_{i(n + 1)} = m
        & \label{eq:constraint2} \\
        & \sum_{i = 0}^{n}
            \phi_{ij} \in [0, 1];
            \ j \in \mathcal{V}'
        & \label{eq:constraint3} \\
        & \sum_{j = 1}^{n + 1}
            \phi_{ij} \in [0, 1];
            \ i \in \mathcal{V}'
        & \label{eq:constraint4} \\
        & \sum_{i = 0}^{n}
            \phi_{ij} = \sum_{i' = 1}^{n + 1}
                \phi_{ji'};
            \ j \in \mathcal{V}'
        & \label{eq:constraint5} \\
        & \sum_{i = 0}^{n}
            \sum_{j = 1}^{n + 1}
                t_{ij} \phi^a_{ij} \leq T;
            \ a \in \mathcal{A}
        & \label{eq:constraint6} \\
        &
            u^a_{i} - u^a_{j} + n\phi^a_{ij} \leq
                n - 1;
            \ a \in \mathcal{A}; \ i, j \in \mathcal{V}'
        & \label{eq:constraint7}
\end{flalign}

In these equations, $|\rho^a|$ denotes the cardinality (number of nodes) of the route $\rho_a$, such that $\rho^a_1=0$ and $\rho^a_{|\rho_a|}=n+1, \forall a \in \mathcal{A}$; the set $\mathcal{V}'=\{1, ..., n\} \subset \mathcal{V}$ represents the non-depot nodes; and $\phi_{ij}=1, \forall i,j \in \mathcal{V}$ if the node $j$ is visited by any agent right after the node $i$, and 0 otherwise. Notice that $\phi_{ij}=1$ applies to all agents while $\phi^a_{ij}=1$ applies to one single agent $a \in \mathcal{A}$.

Equation \ref{eq:goalFunc} is subject to constraints from Equations \ref{eq:constraint1} to \ref{eq:constraint7}, which ensure the feasibility of the TOP solution. Constraints \ref{eq:constraint1} and \ref{eq:constraint2} impose that all agents depart from the start depot (node $i=0$) and arrive at the end depot (node $i=n+1$). Constraints from Equations \ref{eq:constraint3} and \ref{eq:constraint4} prevent any node from being visited more than once, either by the same or different agents. Equation \ref{eq:constraint5} enforces every possible path to be continuous. Restriction from Equation \ref{eq:constraint6} imposes a time limit $T$, where $t_{ij} = d_{ij} / v$ is the time taken by any agent to travel from node $i$ to node $j$. Finally, the subtour elimination constraint \cite{Miller1960} is expressed in Equation \ref{eq:constraint7}, where $u^a_{i}, u^a_{j} = 1, ..., n$ correspond to the positions of nodes $i$ and $j$ in the tour of agent $a$.

\section{System description}
\label{sec:system}

\begin{figure*}[t]
    \centerline{\includegraphics[width=0.9\textwidth]{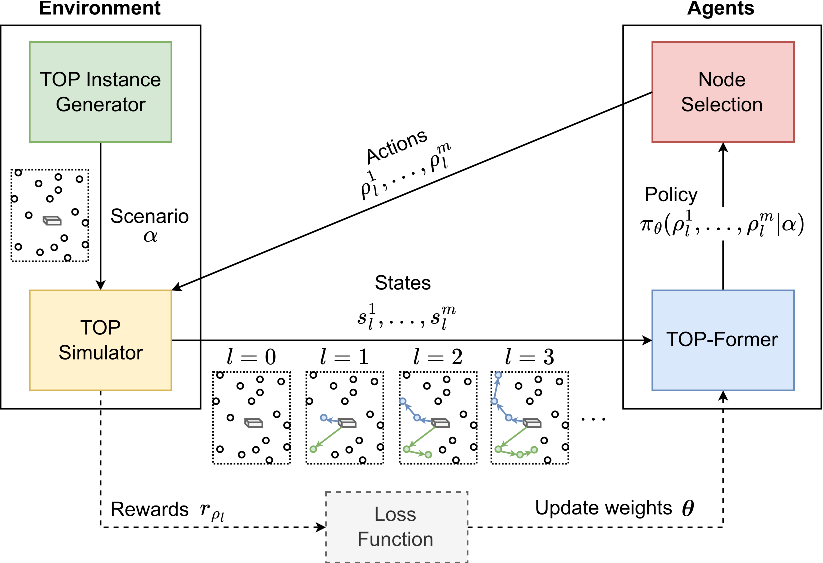}}
    \caption{Training scheme of the proposed TOP-Former, composed of five main blocks: TOP Instance Generator, TOP Simulator, TOP-Former, Node Selection, and Loss Function.}
    \label{fig:scheme}
\end{figure*}

\begin{figure}[t]
    \centerline{\includegraphics[width=0.65\columnwidth]{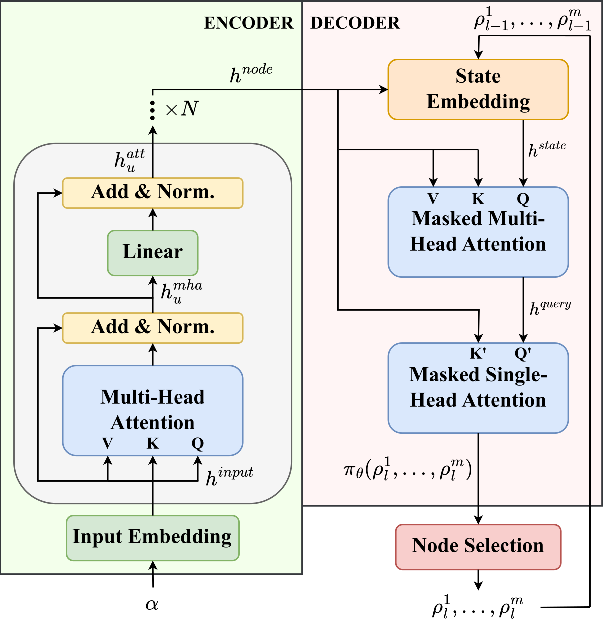}}
    \caption{Structure of TOP-Former, consisting of an Encoder and a Decoder that generate a policy $\pi_{\theta}$ from a given scenario $\alpha$. The final Node Selection module samples the routes $\rho^1, ..., \rho^m$.}
    \label{fig:encoder_decoder}
\end{figure}

TOP-Former network is proposed to solve the TOP, described in Section \ref{sec:problem}. First, the workflow and training strategy are detailed in Section \ref{sec:training}. Next, the architecture of the Transformer-like model is explained in Sections \ref{sec:encoder} and \ref{sec:decoder}, where the two main modules of the network are presented: the Encoder and the Decoder. Finally, the output of the system is described in Section \ref{sec:selection}, which outlines the process of selecting sequences of nodes to visit.

\subsection{Training Strategy}
\label{sec:training}

The training process of the proposed TOP-Former neural network is illustrated in Figure \ref{fig:scheme}, consisting of five main components: TOP Instance Generator, TOP Simulator, TOP-Former, Node Selection, and Loss Function. The input to the TOP-Former Network is a scenario/problem instance $\alpha$ generated by the TOP Simulator, comprising a set of regions to be visited (defined by their coordinates), along with the start and end depots. The network learns a policy $\pi_{\theta} (\rho_1, ..., \rho_m | \alpha)$ that estimates the probability distribution of optimal paths $\rho_1, ..., \rho_m \in \mathcal{P}$ for each agent, where $\theta$ denotes the trained weights of the network. The specific set of predicted paths for each agent, obtained from the learned policy, is determined by the Node Selection component, considering constraints outlined in Section \ref{sec:problem}. Subsequently, the TOP Simulator updates the agents' states iteratively until they all reach the end depot. Since the network learns to find the optimal policy by means of the Reinforce training algorithm \cite{Williams1992}, the final collected reward is employed and combined with Equation \ref{eq:goalFunc} to define the Reinforce loss as:

\vspace*{-4mm}
\begin{gather} \label{eq:lossFunc}
    \mathcal{L(\theta | \alpha)} =
    E_{\pi_{\theta} (\rho^1, ..., \rho^m | \alpha)} \left[
        L(\rho^1, ..., \rho^m)
    \right], \\
    L(\rho_1, ..., \rho_m) =
    - \sum_{a = 1}^{m}
        \sum_{l = 1}^{|\rho^a|-1}
            r_{\rho^a_l} \phi_{\rho^a_l \rho^a_{l+1}}, \nonumber
\end{gather}

where $\mathcal{L(\theta | \alpha)}$ is the loss function, that is, the expectation of the cost function $L(\rho_1, ..., \rho_m)$ following the policy $\pi_{\theta} (\rho_1, ..., \rho_m | \alpha)$. Notice the minus sign indicating that better paths are obtained by minimizing the loss function. To minimize the loss function, gradient descent is applied using the approximation of the gradient given by the Reinforce algorithm, whose expression is:

\vspace*{-5mm}
\begin{gather} \label{eq:gradFunc}
    \nabla \mathcal{L}(\theta | \alpha) \approx
    \left(
        L(\rho_1, ..., \rho_m) - b(\alpha)
    \right) \nabla \log \pi_{\theta} (
        \rho_1, ..., \rho_m | \alpha
    ), \\
    b(\alpha) =
    L \left(
        \argmax_{\rho_1, ..., \rho_m}
            \pi_{\theta} (\rho_1, ..., \rho_m | \alpha)
    \right) \nonumber,
\end{gather}

where $b(\alpha)$ is a baseline that reduces the variance of the estimated gradient, since it represents the greedy route for the instance $\alpha$.

\subsection{Encoder}
\label{sec:encoder}

The policy $\pi_{\theta} (\rho_1, ..., \rho_m | \alpha)$ that predicts the probability distribution of the paths of the agents is estimated by the proposed Transformer-like architecture, which has an Encoder-Decoder structure shown in Figure \ref{fig:encoder_decoder}. While the Encoder structure follows \cite{Fuertes2023}, the Decoder is a centralized evolution to consider the context of every agent for the predictions of multiple simultaneous tours for the TOP.

\begin{figure}[t]
    \centerline{\includegraphics[width=0.4\columnwidth]{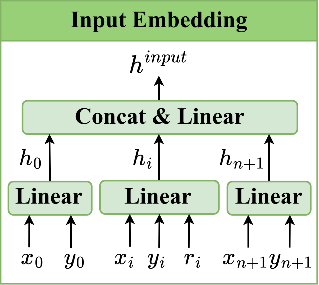}}
    \caption{Structure of the Input Embedding module, which generates an initial graph representation $h^{input}$.}
    \label{fig:input_embed}
\end{figure}

The Encoder estimates a vectorized representation of a given $\alpha$ defined by the coordinates $(x_i, y_i), i \in \mathcal{V}'$ and rewards $r_i, i \in \mathcal{V}'$ of the nodes, the coordinates of the depots $(x_0, y_0)$ and $(x_{n+1}, y_{n+1})$, and the time limit $T$. For this purpose, the Encoder uses an Input Embedding (see Figure \ref{fig:input_embed}) and $N$ encoding blocks (see Figure \ref{fig:encoder_decoder}). The Input Embedding creates an embedding of $\alpha$ in two steps. First, three individual representations are learned (see Equation \ref{eq:inputEmbeddings}): $h_i$, $h_0$, and $h_{n+1}$ for the nodes (coordinates and rewards are projected), and the start and end depots (only coordinates are projected), respectively. The second step computes a global input embedding by concatenating and projecting $h_i$, $h_0$, and $h_{n+1}$ (see Equation \ref{eq:inputEmbedding}), resulting in $h^{input}$, with hidden dimension $\lambda$. Notice that the variables $W$ and $b$ represent the weights and bias of each linear embedding.

\begin{figure}[t]
    \centerline{\includegraphics[width=0.4\columnwidth]{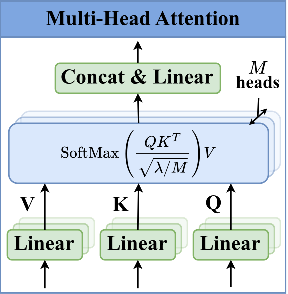}}
    \caption{Multi-Head Attention block, where embeddings $V$, $K$, and $Q$ undergo an attention mechanism, followed by concatenation and a fully connected layer.}
    \label{fig:mha}
\end{figure}

\vspace*{-5mm}
\begin{gather} \label{eq:inputEmbeddings}
    h_0 = W^0\left[x_0, y_0\right] + b^0 \nonumber \\
    h_i = W^i\left[x_i, y_i, r_i\right] + b^i \\
    h^{n+1} = W^{n+1}\left[x_{n+1}, y_{n+1}\right] + b^{n+1} \nonumber
\end{gather}
\begin{gather} \label{eq:inputEmbedding}
    h^{input} = W^{input} \left[ h_0, h_i, h_{n+1} \right] + b^{input},
\end{gather}

The input embedding $h^{input}$ is later processed with $N$ encoding blocks, similar to those proposed in \cite{Vaswani2017}, to find the relations among all the nodes, i.e., to encode the underlying node graph of a given instance problem. The core of every encoding block is a Multi-Head Attention mechanism (see Figure \ref{fig:mha}) that compares and relates the information of every node to the others. As a result, a node embedding ($h^{node}$) containing high-level semantic information about the problem instance and the underlying node graph representation describe the mathematical operation of the encoding blocks. $h^{node}$ can be expressed as

\vspace*{-5mm}
\begin{gather} \label{eq:nodeEmbedding}
    h^{mha}_u = \textrm{BN}\left(h^{mha}_{u-1} + \textrm{MHA}\left(h^{mha}_{u-1}, h^{mha}_{u-1}, h^{mha}_{u-1}\right)\right); u \in {1, ..., N} \\
    h^{att}_u = \textrm{BN}\left(h^{mha}_u + W^{mha}_u h^{mha}_u + b^{mha}_u\right); u \in {1, ..., N} \nonumber \\
    h^{node} = h^{att}_N, \nonumber
\end{gather}

where BN refers to the batch normalization operation, MHA is the Multi-Head Attention mechanism with $M$ heads (see Equation \ref{eq:mha}), and the input layer to the first encoding block is $h^{mha}_{0}=h^{input}$.

\vspace*{-5mm}
\begin{gather}
    Q = W^1 h_1 + b^1 \nonumber \\
    K = W^2 h_2 + b^2 \nonumber \\
    V = W^3 h_3 + b^3 \nonumber \\
    \textrm{MHA}(h_1, h_2, h_3) = \textrm{SoftMax} \left( \frac{QK^T}{\sqrt{\lambda/M}} \right) V
    \label{eq:mha}
\end{gather}

\begin{figure}[t]
    \centerline{\includegraphics[width=0.45\columnwidth]{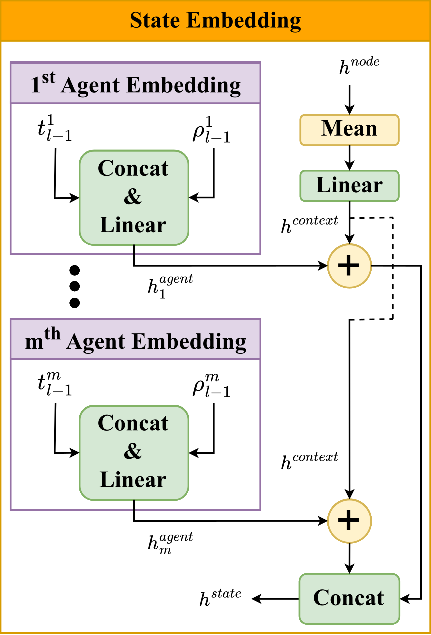}}
    \caption{Structure of the State Embedding module, which integrates node embeddings from the Encoder with agent-specific information (remaining time and previously visited nodes) to produce the state representation $h^{state}$.}
    \label{fig:state_embed}
\end{figure}

\subsection{Decoder}
\label{sec:decoder}

The Decoder considers multiple agents in a simultaneous manner for the TOP and predicts the best policy distribution $\pi_{\theta}$ to conform every agent's path using the node embedding $h^{node}$ provided by the Encoder. The predictions of new nodes in the paths $\{\rho^1_l, ..., \rho^m_l\}$ of the agents are sequentially estimated for each iteration $l \in \{1, ..., |\rho^a|: a \in \mathcal{A}\}$ until the complete tours $\{\rho^1, ..., \rho^m\}$ are found. Hence, the Decoder follows an autoregressive approach, meaning it is executed recursively to generate node predictions until the full sequence is obtained. In contrast, the Encoder computes the node embeddings $h^{node}$ only once. Consequently, to preserve computational efficiency, the Encoder can incorporate $N$ stacked encoding blocks, whereas the Decoder does not.

The Decoder is composed of three main blocks: State Embedding, Masked Multi-Head Attention, and Masked Single-Head Attention. The first block, State Embedding (see Figure \ref{fig:state_embed}), encodes the state of the optimization problem for every iteration, considering the location of the agents, the past visited nodes, the remaining time, and the node embedding $h^{node}$ (with all the scenario information). The computation of the state embedding can be split into 4 steps. The first step computes a new projection from the mean of the node embeddings (see Equation \ref{eq:contextEmbedding}) to have a compact representation of the problem nodes.

\vspace*{-2mm}
\begin{equation} \label{eq:contextEmbedding}
    h^{context} = W^{context}\frac{1}{n} \sum_{i=1}^n h^{node}_i + b^{context}
\end{equation}

The second step calculates the agent embeddings (see Equation \ref{eq:agentEmbedding}), which contain information about the agent state, i. e., the position (coordinates) and the time left for every agent to reach the end depot $t^{a}_l, \forall a \in \mathcal{A}$.

\vspace*{-4mm}
\begin{gather} \label{eq:agentEmbedding}
    t^{a}_l =
    T - \sum_{k=2}^{l}
        t_{\rho^a_k\rho^a_{k-1}}, \
        l \in \{1, ..., \rho^a\} \\
    h^{agent}_a =
    W^{agent}_a \left[
        t^a_l, x_{\rho^k_{l-1}}, y_{\rho^a_{l-1}}
    \right], \nonumber
\end{gather}
\vspace*{-3mm}

The third step aggregates the context embedding $h^{context}$ to every agent embedding $h^{agent}_a$. Finally, the output of each previous aggregation is concatenated (see Equation \ref{eq:queryEmbedding}) generating the final state representation $h^{state}$, which will be the query for the following Masked Multi-Head Attention module. This query will contain information about the state of the agents and their context.

\vspace*{-4mm}
\begin{equation} \label{eq:queryEmbedding}
    h^{state} = \left[h^{context} + h^{agent}_1, ..., h^{context} + h^{agent}_m\right]
\end{equation}

The Masked Multi-Head Attention block relates every node of the problem instance $\alpha$ with the state of every agent. The key and value inputs are projected from the Encoder's node embeddings $h_{node}$, while the query is obtained from the previous State Embedding block. The involved operations are very similar to the previously described Multi-Head Attention block from the Encoder, but taking into account the masked procedure $f_{mask}$.

\vspace*{-2.5mm}
\begin{equation} \label{eq:mask}
    f_{mask} (h_i) = 
    \begin{cases}
        h_i     & \text{ if } i \in \mathcal{V}^l \\
        -\infty & \text{otherwise}
    \end{cases},
    i \in \mathcal{V}'
\end{equation}

Here, $\mathcal{V}^l$ represents the set of nodes that can be visited during the iteration $l \in \{1, ..., |\rho^a|: a \in A\}$. Notice that $\mathcal{V}^l$ is designed to comply with constraints from Section \ref{sec:problem}. Thus, we impose hard constraints by fixing the probability of prohibited nodes to $0$. This includes already-visited nodes (see Equations \ref{eq:constraint3}, \ref{eq:constraint4}, \ref{eq:constraint5}, and \ref{eq:constraint7}), and nodes that cannot be visited on time after reaching the end depot (see Equations \ref{eq:constraint2} and \ref{eq:constraint6}). Besides, we comply with Equation \ref{eq:constraint1} by initializing the agents' state at the start depot location.

The resulting output of the Masked Multi-Head Attention block, which will be the new query $h^{query}$ for the next block, will contain enough discriminating information to predict the best next node for every agent. The following Masked Single-Head Attention applies the Attention mechanism again (without multiple heads) to obtain the log-probabilities or logits (non-normalized probabilities) $\pi_{\theta}^{logits}$ of the new nodes to be aggregated to the agents' routes. Its mathematical computation is given by Equation \ref{eq:logits}.

\begin{equation} \label{eq:logits}
    \pi_{\theta}^{logits} (\rho^1, ..., \rho^m | \alpha) = C \cdot \textrm{tanh} \left( \frac{Q'K'^T}{\sqrt{\lambda}} \right),
\end{equation}

where the constant $C$ and the tanh clipping function control the range of the logits, and $Q'$ and $K'$ are projections of query and key inputs ($h^{query}$ and $h^{node}$), respectively. The logits are later normalized by applying a SoftMax operator to obtain the final policy $\pi_{\theta} (\rho^1, ..., \rho^m | \alpha)$.

\subsection{Node Selection}
\label{sec:selection}

Lastly, the selection of the nodes is drawn from the policy $\pi_{\theta}$ by means of two different strategies. During training, an exploration versus exploitation strategy is desired to reinforce good decisions and explore potentially better decisions. During inference, and also at the time of calculating the baseline $b(\alpha)$ from Equation \ref{eq:gradFunc}, a greedy selection (exploitation) is recommended to find the best routes.

There is an additional difficulty in the node selection process for the multi-agent case due to the possibility of two or more agents choosing the same node at the same time. To avoid this case, decisions are chosen in order for every agent in such a way that each agent selection blocks the selected nodes for the rest of the agents that have not taken their action yet. The SoftMax operator is applied after each agent's node selection to re-normalize the probabilities, compensating the effect of the blocking method, which consists in forcing the value of the chosen node to $-\infty$ (similar to Equation \ref{eq:mask}).

\section{Results}
\label{sec:results}

The set of experiments performed to validate and compare the proposed system with other algorithms from the state-of-the-art is presented in this Section. More specifically, the experimental setup is explained in section \ref{sec:setup}, while the results, discussions, and comparisons with other methods are detailed in Section \ref{sec:discussions}. Section \ref{sec:issues} introduces some open issues for the TOP.

\subsection{Experimental setup}
\label{sec:setup}

\begin{figure*}[t]
    \centerline{\includegraphics[width=\textwidth]{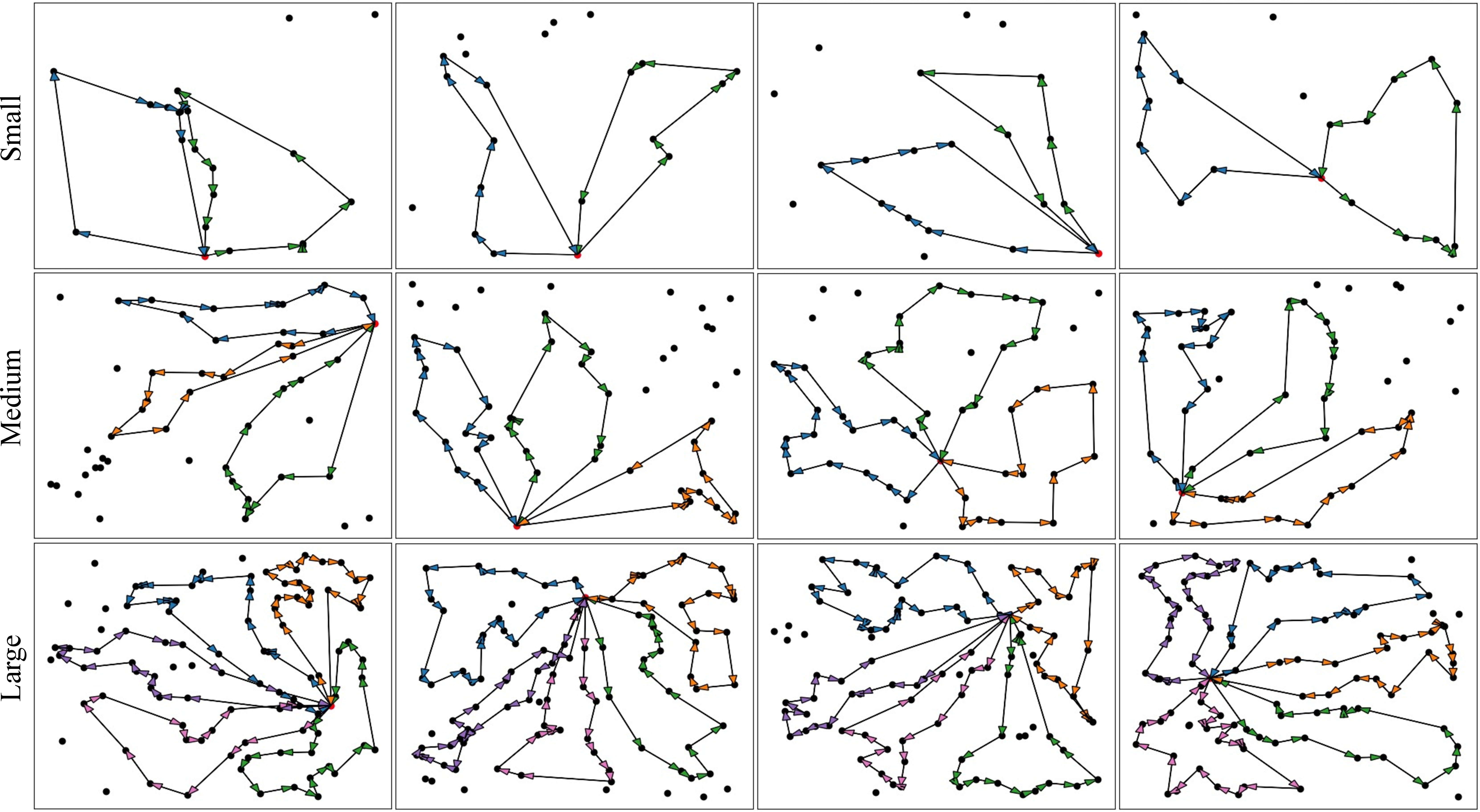}}
    \caption{Qualitative results of TOP-Former on synthetic scenarios of different sizes: small ($n=20$, $m=2$), medium ($n=50$, $m=3$), and large ($n=100$, $m=5$). Black circles represent nodes, while red circles denote depots. Tours are color-coded per agent.}
    \label{fig:qualitative}
\end{figure*}

\begin{figure*}[t]
    \centerline{\includegraphics[width=0.85\textwidth]{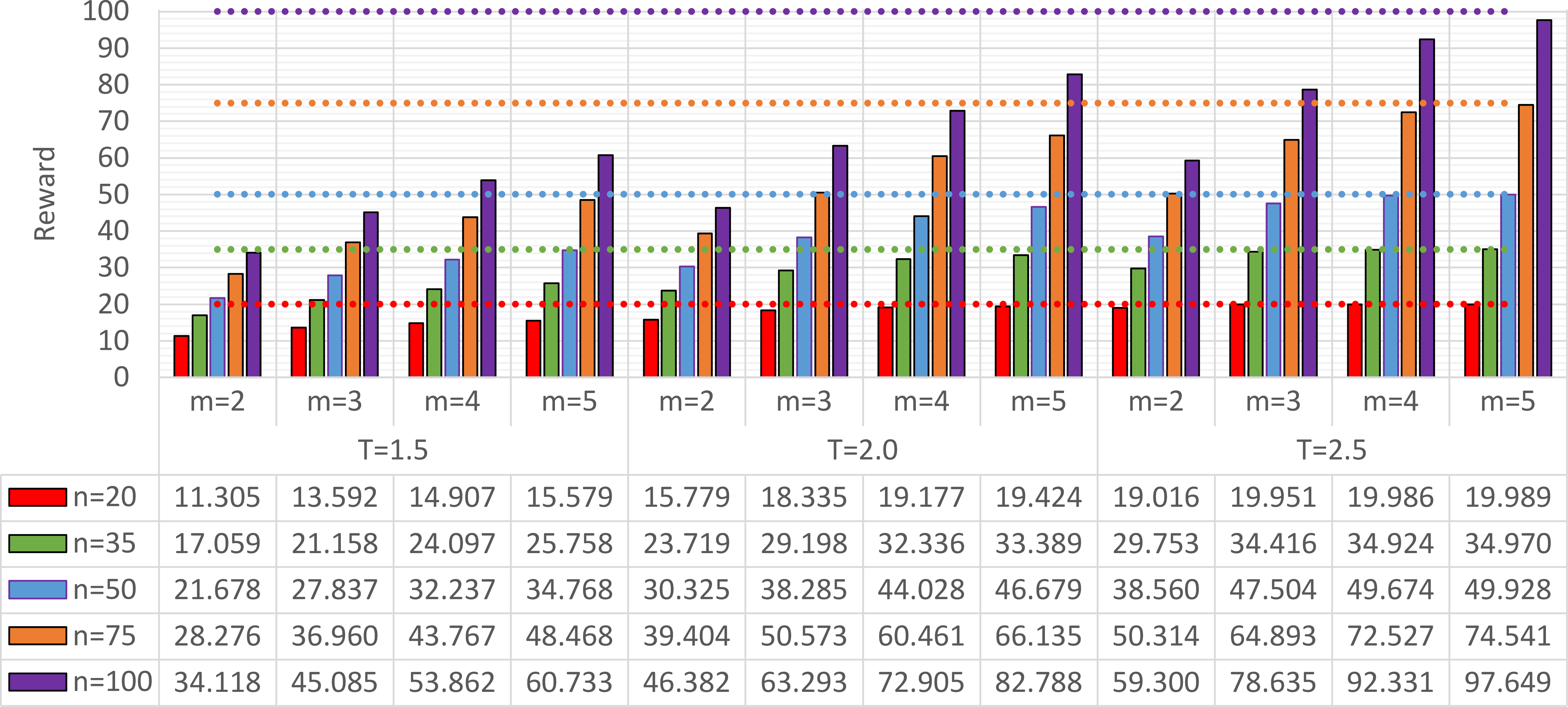}}
    \caption{Reward collection performance of TOP-Former across various scenarios, considering different values of $n$, $m$, and $T$, with constant rewards $r_i=1$. The maximum collectable rewards are indicated by dashed lines.}
    \label{fig:results}
\end{figure*}

To evaluate TOP-Former, both synthetic and real-world datasets were used. For the synthetic dataset, several instance problems were generated to train, validate, and test the proposed system. The test and validation datasets contain 10,000 instances, whereas the training data is composed of more than a million instances that are randomly generated during every training epoch. Each instance problem is determined by the depot location, the set of regions to visit, and the time given to visit the regions and return to the depot. The coordinates of the regions and the depot are normalized to the range [0, 1] to improve network comprehension, ensuring that $x_i, y_i \in [0, 1], i \in \mathcal{V}$. The number of depots is one, which means that every agent must start and end their tours at the same location (i.e., $x_0=x_{n+1}; y_0=y_{n+1}$). The reward associated with each region is fixed at three different values: a constant value of 1 ($r_i=1, \forall i \in \mathcal{V}'$), a reward sampled from the uniform distribution ($r_i \sim \mathcal{U}(0.01, 1)$), and a reward value based on the distance to the depot, as proposed in \cite{Fischetti1998}. Different type of reward values may indicate preference of visiting some nodes over others.

For the experiments, multiple types of instance problems are considered, including scenarios with $n = \{ 20, 35,$ $50, 75, 100 \}$ nodes solved by fleets of $m = \{ 2, 3, 4, 5 \}$ agents with time limits of $T = \{ 1.5, 2, 2.5 \}$ (the set of values are expanded versions from those used in \cite{Kool2019,Bello2017,Fuertes2023,Sankaran2022}). This allows for studying the performance of the system under different environmental conditions. However, for comparison with other state-of-the-art methods, we considered three types of instance problems, depending on the number of feasible solutions that can be found: small ($n=20$, $m=2$), medium ($n \in \{40,60\}$, $m=3$), and large ($n=100$, $m=5$). This categorization allows for evaluating the scalability of each algorithm with respect to the number of agents and nodes. Moreover, the time limit is fixed to $T=2$ since it represents the most challenging case for the TOP, as explained in \cite{Vansteenwegen2009}.

Regarding the real-world data, the considered dataset is VDRPMDPC (Van Drone Routing Problem with Multiple Delivery Points and Cooperation) \cite{Athanasiadis2023}, and includes $14~$instances of TOP scenarios (consisting of a depot and a set of clients) for a package delivery application. Due to the limited number of instances, data augmentation was applied. First, the coordinates were normalized to the range [0, 1], similar to the synthetic dataset. Then, Gaussian noise with a standard deviation of 0.01 was added to each coordinate. After generating 35 augmentations per original instance, the final dataset contained 504 instances, including the original samples. The dataset contains instances of 20, 40, 60, and 100 nodes. We followed the experimental setup explained for the synthetic dataset, resulting in: small ($n=20$, $m=2$), medium ($n \in \{40,60\}$, $m=3$), and large ($n=100$, $m=5$) scenarios. Additionally, in this case, the time limit represents the service time available for the couriers (the agents) to deliver as many packages as possible. For the experiments, this time limit is fixed to $T=2$, as it was explained for the synthetic dataset. Moreover, the reward of each region was set to a constant value of 1.

As accuracy metrics, we consider the average collected reward and average number of nodes visited from the solutions of the test instances. Additionally, we include the gap with respect to the highest reward value to make a direct comparison between algorithms. The gap is calculated as follows:

\begin{equation} \label{eq:gap}
    \text{Gap}_k (\%) = \left( 1 - \frac{R_k}{\max_{\hat{k}}R_{\hat{k}}} \right) * 100
\end{equation}

Here, $\text{Gap}_k$ is the gap value of algorithm $k$ with respect to the best, and $R_k$ is the reward collected by algorithm $k$. Notice that, for the best algorithm in terms of collected reward, the gap value is 0\%.

Regarding the hyperparameter tuning, the Encoder is composed of $N=3$ consecutive encoding blocks with a hidden dimension of $\lambda=128$ that is preserved along the entire network. Since the batch size applied during the training stage is the largest allowed by the GPU memory, batch normalization is applied by the Encoder to control the gradient instead of layer normalization. The range of the logits generated by the Masked Single-Head Attention of the Decoder is controlled by the typical constant value $C=10$. To train the network, the Adam optimizer is used in combination with a learning rate of $10^{-4}$ along 100 epochs. Two NVIDIA Titan Xp GPUs are used to speed up the training process. The CPU is an Intel Core i9-7900X 3.30GHz. The inference time related to the proposed and the state-of-the-art methods has been calculated on both CPU and GPU (if possible). The code and the datasets used are publicly available \footnote{\url{http://gti.ssr.upm.es/data}}.

\subsection{Results and discussions}
\label{sec:discussions}

\begin{table*}[t]
    \caption{Comparison between state-of-the-art methods (best in bold) on synthetic scenarios with constant reward values per node. Comparison is in terms of average collected reward, gap percentage (with respect to highest reward), and average number of nodes visited.}
    \label{tab:comparison-const}
    \centering
    \begin{tabular}{r|rrr|rrr|rrr}
        \multirow{2}{*}{Reward type: Constant} & \multicolumn{3}{c|}{Small} & \multicolumn{3}{c|}{Medium} & \multicolumn{3}{c}{Large} \\
        & Reward & Gap (\%) & Nodes & Reward & Gap (\%) & Nodes & Reward & Gap (\%) & Nodes \\
        \hline\hline
        TOP-Former & 15.779 & 0.196 & 15.779 & 38.295 & 0.083 & 38.295 & \textbf{82.788} & \textbf{0.000} & \textbf{82.788} \\
        PN \cite{Bello2017} & 14.512 & 8.210 & 14.512 & 36.094 & 5.826 & 36.094 & 80.214 & 3.109 & 80.214 \\
        GPN \cite{Ma2020} & 14.625 & 7.495 & 14.625 & 36.315 & 5.250 & 36.315 & 78.873 & 4.729 & 78.873 \\
        GAMMA \cite{Sankaran2022} & 15.726 & 0.531 & 15.726 & 38.221 & 0.277 & 38.221 & 81.108 & 2.029 & 81.108 \\
        GA \cite{Xiao2022} & 13.280 & 16.003 & 13.280 & 24.924 & 34.970 & 24.924 & 36.082 & 56.416 & 36.082 \\
        PSO \cite{Xiao2022} & 15.798 & 0.076 & 15.798 & 16.905 & 55.893 & 16.905 & 25.442 & 69.268 & 25.442 \\
        ACO \cite{Xiao2022} & 15.800 & 0.063 & 15.800 & \textbf{38.327} & \textbf{0.000} & \textbf{38.327} & 81.387 & 1.692 & 81.387 \\
        Gurobi (15s) \cite{Gurobi2024} & 14.982 & 5.237 & 14.982 & 19.034 & 50.338 & 19.034 & 9.618 & 88.382 & 9.618 \\
        Gurobi (30s) \cite{Gurobi2024} & 15.432 & 2.391 & 15.432 & 19.682 & 48.647 & 19.682 & 16.782 & 79.729 & 16.782 \\
        Gurobi (60s) \cite{Gurobi2024} & \textbf{15.810} & \textbf{0.000} & \textbf{15.810} & 21.547 & 43.781 & 21.547 & 31.535 & 61.909 & 31.535 \\
    \end{tabular}
\end{table*}

\begin{table*}[t]
    \caption{Comparison between state-of-the-art methods (best in bold) on synthetic scenarios with reward values sampled from uniform distribution. Comparison is in terms of average collected reward, gap percentage (with respect to highest reward), and average number of nodes visited.}
    \label{tab:comparison-unif}
    \centering
    \begin{tabular}{r|rrr|rrr|rrr}
        \multirow{2}{*}{Reward type: Uniform} & \multicolumn{3}{c|}{Small} & \multicolumn{3}{c|}{Medium} & \multicolumn{3}{c}{Large} \\
        & Reward & Gap (\%) & Nodes & Reward & Gap (\%) & Nodes & Reward & Gap (\%) & Nodes \\
        \hline\hline
        TOP-Former & 8.266 & 1.771 & 14.841 & \textbf{20.268} & \textbf{0.000} & 35.296 & \textbf{43.309} & \textbf{0.000} & \textbf{76.842} \\
        PN \cite{Bello2017} & 7.662 & 8.948 & 13.352 & 18.613 & 8.166 & 32.017 & 40.157 & 7.278 & 70.047 \\
        GPN \cite{Ma2020} & 7.704 & 8.449 & 13.482 & 19.035 & 6.083 & 33.362 & 41.250 & 4.754 & 73.348 \\
        GAMMA \cite{Sankaran2022} & 8.241 & 2.068 & 14.690 & 20.170 & 0.484 & \textbf{35.702} & 42.257 & 2.429 & 74.750 \\
        GA \cite{Xiao2022} & 7.428 & 11.729 & 12.064 & 14.616 & 27.886 & 21.261 & 22.854 & 47.230 & 31.654 \\
        PSO \cite{Xiao2022} & 8.225 & 2.258 & 13.641 & 10.371 & 48.831 & 15.511 & 15.376 & 64.497 & 23.222 \\
        ACO \cite{Xiao2022} & 8.223 & 2.282 & 13.638 & 19.609 & 3.251 & 31.628 & 41.029 & 5.264 & 67.088 \\
        Gurobi (15s) \cite{Gurobi2024} & 7.996 & 4.979 & 14.396 & 10.594 & 47.730 & 17.909 & 4.850 & 88.801 & 9.539 \\
        Gurobi (30s) \cite{Gurobi2024} & 8.229 & 2.210 & 14.951 & 10.954 & 45.954 & 18.486 & 8.675 & 79.970 & 15.390 \\
        Gurobi (60s) \cite{Gurobi2024} & \textbf{8.415} & \textbf{0.000} & \textbf{15.391} & 11.969 & 40.946 & 20.083 & 18.359 & 57.609 & 29.434 \\
    \end{tabular}
\end{table*}

\begin{table*}[!t]
    \caption{Comparison between state-of-the-art methods (best in bold) on synthetic scenarios with reward values based on distance to depot. Comparison is in terms of average collected reward, gap percentage (with respect to highest reward), and average number of nodes visited.}
    \label{tab:comparison-dist}
    \centering
    \begin{tabular}{r|rrr|rrr|rrr}
        \multirow{2}{*}{Reward type: Distance} & \multicolumn{3}{c|}{Small} & \multicolumn{3}{c|}{Medium} & \multicolumn{3}{c}{Large} \\
        & Reward & Gap (\%) & Nodes & Reward & Gap (\%) & Nodes & Reward & Gap (\%) & Nodes \\
        \hline\hline
        TOP-Former & 8.527 & 1.422 & 15.231 & \textbf{19.572} & \textbf{0.000} & \textbf{36.746} & \textbf{41.256} & \textbf{0.000} & 78.601 \\
        PN \cite{Bello2017} & 8.068 & 6.728 & 14.324 & 16.422 & 16.094 & 31.131 & 27.750 & 32.737 & 47.947 \\
        GPN \cite{Ma2020} & 7.986 & 7.676 & 14.284 & 17.951 & 8.282 & 33.789 & 38.126 & 7.587 & 73.511 \\
        GAMMA \cite{Sankaran2022} & 8.494 & 1.803 & 15.093 & 19.476 & 0.490 & 36.561 & 40.657 & 1.452 & \textbf{78.794} \\
        GA \cite{Xiao2022} & 7.334 & 15.214 & 12.410 & 13.151 & 32.807 & 22.162 & 19.628 & 52.424 & 31.961 \\
        PSO \cite{Xiao2022} & \textbf{8.650} & \textbf{0.000} & 15.176 & 8.868 & 54.690 & 15.236 & 12.534 & 69.619 & 22.522 \\
        ACO \cite{Xiao2022} & \textbf{8.650} & \textbf{0.000} & 15.194 & 19.064 & 2.596 & 34.939 & 37.995 & 7.904 & 72.693 \\
        Gurobi (15s) \cite{Gurobi2024} & 8.054 & 6.890 & 14.501 & 8.795 & 55.063 & 15.065 & 4.650 & 88.729 & 15.012 \\
        Gurobi (30s) \cite{Gurobi2024} & 8.380 & 3.121 & 15.082 & 9.132 & 53.342 & 17.897 & 8.344 & 79.775 & 16.772 \\
        Gurobi (60s) \cite{Gurobi2024} & 8.620 & 0.347 & \textbf{15.491} & 10.088 & 48.457 & 19.441 & 14.778 & 64.180 & 28.510 \\
    \end{tabular}
\end{table*}

Figure \ref{fig:qualitative} contains qualitative visual examples of TOP solutions predicted by the TOP-Former network. Examples of randomly-generated synthetic scenarios of small, medium, and large size are displayed to show the versatility of the network for any kind of situation. It can be observed that the tours proposed are qualitatively good since few or no redundant paths are generated.

Additionally, quantitative results considering the multiple environmental conditions (mentioned in Section \ref{sec:setup}) have been performed. Figure \ref{fig:results} exposes the performance of the system with different types of synthetic scenarios (changing $n, m,$ and $T$) and constant rewards ($r_i=1$). As expected, while $m$ and $T$ grow, the reward collection is higher and saturates for extreme situations, such as the case with $m=5$ and $T=2.5$, where almost all nodes can be visited. Notice also that, for the same values of $m$ and $T$, the reward grows as $n$ is increased. The reason is that the density of regions contained on the same map (normalized square of coordinates in the range [0, 1]) is higher for larger values of $n$, which allows increasing the number of node visits for the same $m$ and $T$.

Quantitative results are expanded by comparing multiple state-of-the-art works for small, medium, and large-sized scenarios with the reward value distributions mentioned in Section \ref{sec:setup} (constant values, values sampled from the uniform distribution, and values based on the distance to the depot). More specifically, the proposed TOP-Former has been compared in terms of accuracy (see Tables \ref{tab:comparison-const}, \ref{tab:comparison-unif}, and \ref{tab:comparison-dist}), and computational time (see Table \ref{tab:time}) with some state-of-the-art methods commented in Section \ref{sec:sota}: linear programming solvers, like Gurobi \cite{Gurobi2024}; heuristic algorithms, such as GA, PSO, and ACO \cite{Xiao2022}; RNN-based networks, like PN \cite{Bello2017} and GPN \cite{Ma2020}; and the Transformer-based neural network GAMMA \cite{Sankaran2022}. It is important to note that Gurobi generates solutions with an impractically long computation time, requiring the use of a specific timeout (15, 30, and 60 seconds per instance problem) and the selection of the best solution found within that limit. Moreover, multiprocessing with 20 CPU cores is applied to the non-neural network methods (Gurobi, GA, PSO, and ACO) to reduce the large computation time required.

From Tables \ref{tab:comparison-const}, \ref{tab:comparison-unif}, and \ref{tab:comparison-dist}, it can be observed that TOP-Former obtains competitive performance on synthetic samples both for reward collection and number of nodes visited, which are directly related. The performance of all the algorithms is similar for each type of reward. Notice that the node visits are significantly higher for constant rewards, which is logical since no node is preferred over the others; hence, no region is sacrificed to reach another one with higher reward.

For small scenarios, Gurobi (60s), PSO, and ACO achieve the highest accuracy, but only ACO is capable of maintaining this performance for medium and large problem instances and it is slightly surpassed by TOP-Former. Moreover, the computational time of Gurobi, PSO, and ACO is prohibitive for real-time applications, as shown in Table \ref{tab:time}. Therefore, it can be stated that TOP-Former scales better in terms of both accuracy and computation time with larger-sized scenarios.

The results of GAMMA are slightly worse than those of TOP-Former, mainly due to its sequential nature, which explains its longer computation time. Additionally, the local context analysis of GAMMA for each agent is more restrictive than the global analysis performed by the proposed network, which explains the superiority of the latter in terms of reward. This global context analysis becomes more crucial for larger scenarios, as it allows for the discovery of better solutions. As a result, TOP-Former is more robust in complex situations and outperforms GAMMA in large-sized scenarios.

As for the other algorithms, GA takes excessive time to achieve very low reward values, making it a non-competitive alternative. On the other hand, the RNN-based networks, PN and GPN, produce good results in terms of reward and computation time but are not sufficient to outperform TOP-Former. This was expected, as Transformers were originally designed to surpass RNNs in both accuracy and computation speed.

The performance analysis is extended to the VDRPMDPC dataset, as shown in Table \ref{tab:VDRPMDPC}. Since constant rewards are considered, the average number of nodes visited is equal to the average collected reward, eliminating the need to report the number of nodes visited separately in Table \ref{tab:VDRPMDPC}. From the table, it can be observed that TOP-Former maintains its competitiveness in terms of reward collection, achieving the highest results for small scenarios and the second-best results for large ones. Although ACO is the most successful method for medium and large problem instances, Table \ref{tab:time} demonstrated that TOP-Former is three orders of magnitude faster. Hence, it can be concluded that TOP-Former achieves the best overall performance among real-time algorithms. This strong performance proves TOP-Former to be suitable for package delivery applications, where fast and efficient decision-making is important. The significant reduction in computation time compared to other methods, along with maintaining or even improving solution quality, allows TOP-Former to optimize the routes of autonomous vehicles in real-time, improving delivery efficiency and reducing operational costs.

\begin{table*}[t]
    \caption{Average computational time (in milliseconds) required by TOP-Former and other state-of-the-art methods for solving TOP instances on both CPU and GPU.}
    \label{tab:time}
    
    \centering
    \begin{tabular}{r|rr|rr|rr}
        & \multicolumn{2}{c|}{Small} &\multicolumn{2}{c|}{Medium} & \multicolumn{2}{c}{Large} \tabularnewline
        Time (ms) & CPU & GPU & CPU & GPU & CPU & GPU \tabularnewline
        \hline
        TOP-Former & \textbf{0.289} & \textbf{0.049} & \textbf{1.124} & \textbf{0.101} & \textbf{3.947} & \textbf{0.221} \tabularnewline
        PN \cite{Bello2017} & 1.259 & 0.097 & 5.862 & 0.462 & 31.539 & 1.913 \tabularnewline
        GPN \cite{Ma2020} & 0.708 & 0.105 & 3.533 & 0.359 & 17.275 & 1.125 \tabularnewline
        GAMMA \cite{Sankaran2022} & 0.316 & 0.054 & 1.360 & 0.137 & 6.937 & 0.387 \tabularnewline
        GA \cite{Xiao2022} & 536.228 & - & 2,305.039 & - & 5,793.142 & - \tabularnewline
        PSO \cite{Xiao2022} & 338.220 & - & 4,386.574 & - & 33,583.184 & - \tabularnewline
        ACO \cite{Xiao2022} & 640.444 & - & 3,652.111 & - & 14,913.103 & - \tabularnewline
        Gurobi (15s) \cite{Gurobi2024} & 750.618 & - & 754.587 & - & 763.474 & - \\
        Gurobi (30s) \cite{Gurobi2024} & 1,501.322 & - & 1,507.543 & - & 1,518.583 & - \\
        Gurobi (60s) \cite{Gurobi2024} & 3,001.279 & - & 3,007.085 & - & 3,007.264 & -
    \end{tabular}
\end{table*}

\begin{table*}[!t]
    \caption{Comparison between state-of-the-art methods (best in bold) on the VDRPMDPC dataset with constant reward values per node. Comparison is in terms of average collected reward and gap percentage (with respect to highest reward).}
    \label{tab:VDRPMDPC}
    \centering

    \begin{tabular}{r|rr|rr|rr}
        \multirow{2}{*}{VDRPMDPC Dataset \cite{Athanasiadis2023}} &
        \multicolumn{2}{c|}{Small} &
        \multicolumn{2}{c|}{Medium} &
        \multicolumn{2}{c}{Large}  \\

        &
        Reward &
        Gap (\%) &
        Reward &
        Gap (\%) &
        Reward &
        Gap (\%) \\
        \hline\hline

        TOP-Former &
        \textbf{16.458} &
        \textbf{0.000} &
        34.889 &
        12.053 &
        90.917 &
        2.079 \\

        PN \cite{Bello2017} &
        15.236 &
        7.425 &
        33.910 &
        14.521 &
        85.688 &
        7.711 \\

        GPN \cite{Ma2020} &
        15.681 &
        4.721 &
        33.112 &
        16.534 &
        85.861 &
        7.524 \\

        GAMMA \cite{Sankaran2022} &
        16.028 &
        2.613 &
        36.181 &
        8.797 &
        88.222 &
        4.981 \\

        GA \cite{Xiao2022} &
        15.097 &
        8.270 &
        27.851 &
        29.795 &
        47.292 &
        49.065 \\

        PSO \cite{Xiao2022} &
        13.792 &
        16.199 &
        24.167 &
        39.082 &
        55.757 &
        39.947 \\

        ACO \cite{Xiao2022} &
        16.097 &
        2.193 &
        \textbf{39.671} &
        \textbf{0.000} &
        \textbf{92.847} &
        \textbf{0.000} \\

        Gurobi (15s) \cite{Gurobi2024} &
        15.444 &
        6.161 &
        21.375 &
        46.119 &
        14.326 &
        84.570 \\

        Gurobi (30s) \cite{Gurobi2024} &
        15.778 &
        4.132 &
        22.570 &
        43.108 &
        31.938 &
        65.601 \\

        Gurobi (60s) \cite{Gurobi2024} &
        16.028 &
        2.613 &
        24.271 &
        38.819 &
        41.444 &
        55.363 \\
        
    \end{tabular}
\end{table*}

\subsection{Open issues}
\label{sec:issues}

An inherent challenge shared by all the studied algorithms is their limited scalability due to their centralized nature. Scalability relates the exponential expansion of the solution space to the number of agents and/or the number of nodes. As the number of agents increases, the simultaneous inference of all routes leads to higher computational costs and, in many cases, a deterioration in accuracy due to the complexity of coordinating multiple decision-making processes. That is, the decision of each agent affects the optimal choices for all other agents. Similarly, increasing the number of nodes results in a larger feasible solution space, requiring additional computational resources.

For learning-based methods, in particular, there are additional challenges in the training process. The first one is related to the exploration versus exploitation strategy (see Section \ref{sec:selection}), since it becomes difficult to balance exploring new potential solutions while exploiting the known (partially-good) ones when the solution space is enormous. The second problem is due to the memory usage required during training (commonly, in GPU), as handling larger problem instances demands higher-capacity models, which may not be feasible without substantial computational infrastructure.

In any case, the scalability limitations are evident across all methods considered in this study. Heuristic approaches, such as ACO and PSO, perform well in small-scale instances due to their iterative nature, which allows fine-tuned solutions. However, as the problem size increases, these methods experience severe performance degradation. The reason is related to the number of iterations required to reach high-quality solutions, which grow significantly and leads to impractical computational times (see Table \ref{tab:time}). Exact solvers, such as Gurobi, face even more severe scalability issues, as the combinatorial nature of the problem renders large instances computationally intractable. On the other hand, RNN-based approaches, including PN and GPN, offer advantages in terms of inference speed, but they struggle with long-range dependencies, limiting their ability to generalize effectively to large problem instances. GAMMA, as a sequential learning approach, processes each agent's route individually, which significantly hinders its capacity to capture global dependencies, making it slightly less efficient than the proposed TOP-Former.

TOP-Former demonstrates superior performance in balancing solution quality and computational efficiency, but it is not exempt from scalability problems. While its self-attention mechanism allows for effective global information processing, the complexity of attention layers makes it computationally expensive for large-scale instances. Specifically, it has a cost of $O(n^2\times d)$, where $n$ is the number of nodes, and $d$ is the hidden dimension of the query, key, and value vector embeddings. Therefore, the complexity is quadratic in the number of nodes. Moreover, the model's centralized nature poses additional challenges when the number of agents increases, as commented before. In any case, its scalability behavior is significantly better than the other approaches.

Future research should explore strategies to mitigate these scalability challenges. Decentralized approaches, where multiple smaller models operate independently while exchanging critical information, could help distribute computational complexity more efficiently. However, such methods must address the inherent trade-off between computational cost and accuracy, as the local nature of decentralized analysis may lead to suboptimal global solutions.

\section{Conclusion}
\label{sec:conclusion}

A centralized Transformer-based neural network capable of effectively solving the Team Orienteering Problem (TOP) is proposed in this paper. The neural network, called TOP-Former, consists of an Encoder-Decoder architecture, where the Encoder examines and encodes the scenario, and the Decoder analyzes the situation of every agent at every time step and simultaneously provides a set of tours that solve the TOP with optimal reward collection. Contrary to other methods, TOP-Former solves the problem in a direct and holistic manner by considering the global context of all agents and acting in consequence. The experiments and comparisons have confirmed that the proposed method is faster and more accurate than other state-of-the-art algorithms. TOP-Former has also a better scalability capability to several scenario parameters like the number of nodes and the number of agents. Future research lines could focus on greatly increasing the number of nodes and agents, since centralized solutions like the proposed Transformer and other state-of-the-art methods for the TOP have that intrinsic disadvantage.

\section*{Acknowledgments}
This work has been partially supported by project PID2023-148922OA-I00 (EEVOCATIONS) funded by MCIU/AEI/10.13039/501100011033 of the Spanish Government, and by project TEC-2024/COM-322 (IDEALCV-CM) funded by Comunidad de Madrid.

\bibliographystyle{unsrt}  
\bibliography{arxiv}  

\end{document}